\documentclass[a4paper, 10pt, conference]{ieeeconf}      
\usepackage{FG2020}

\FGfinalcopy 

\IEEEoverridecommandlockouts                              
\usepackage{graphics} 
\usepackage{epsfig} 
\usepackage{amsmath} 
\usepackage{amssymb}  
\usepackage{multirow}
\usepackage{dblfloatfix} 
\usepackage{subcaption}
\usepackage{pifont}
\usepackage{mathtools}
\usepackage{booktabs}

\newcommand{\xmark}{\ding{55}}
\newcommand{\norm}[1]{\left\lVert#1\right\rVert}

\newcommand{\mytensor}[1]{\ensuremath{\mathcal{#1}}}
\newcommand{\myvector}[1]{\ensuremath{\mathbf{#1}}}

\title{\LARGE \bf
Toward fast and accurate human pose estimation via soft-gated skip connections
}

\author{\parbox{16cm}{\centering
    {\large Adrian Bulat, Jean Kossaifi, Georgios Tzimiropoulos, Maja Pantic}\\
    {\normalsize
    Samsung AI Center, Cambridge, UK\\
    adrian@adrianbulat.com, jean.kossaifi@gmail.com, georgios.t@samsung.com, maja.pantic@samsung.com}}
}

\begin{document}

\ifFGfinal
\thispagestyle{empty}
\pagestyle{empty}
\else
\author{Anonymous FG2020 submission\\ Paper ID \FGPaperID \\}
\pagestyle{plain}
\fi
\maketitle

\begin{abstract}

This paper is on highly accurate and highly efficient human pose estimation. Recent works based on Fully Convolutional Networks (FCNs) have demonstrated excellent results for this difficult problem. While residual connections within FCNs have proved to be quintessential for achieving high accuracy, we re-analyze this design choice in the context of improving \textit{both the accuracy and the efficiency} over the state-of-the-art. In particular, we make the following contributions: (a) We propose gated skip connections with per-channel learnable parameters to control the data flow for each channel within the module within the macro-module. (b) We introduce a hybrid network that combines the HourGlass and U-Net architectures which minimizes the number of identity connections within the network and increases the performance for the same parameter budget. Our model achieves state-of-the-art results on the MPII and LSP datasets. In addition, with a reduction of $3\times$ in model size and complexity, we show no decrease in performance when compared to the original HourGlass network.

\end{abstract}

\section{Introduction}\label{sec:introduction}

Being one of the most challenging computer vision problems with a multitude of applications, human pose estimation has been one of the primary research areas that the computer vision community tried to solve with Deep Learning and Convolutional Neural Networks (CNNs). Given that the results produced by existing state-of-the-art methods look at least impressive both qualitatively and quantitatively, it is natural to question how much progress can be expected on this problem over the next years and whether there is room for further improvement. 

\begin{figure}[t]
\centering
\includegraphics[trim={0.5cm 0.5cm 0.5cm 0.5cm},clip,width=7.0cm]{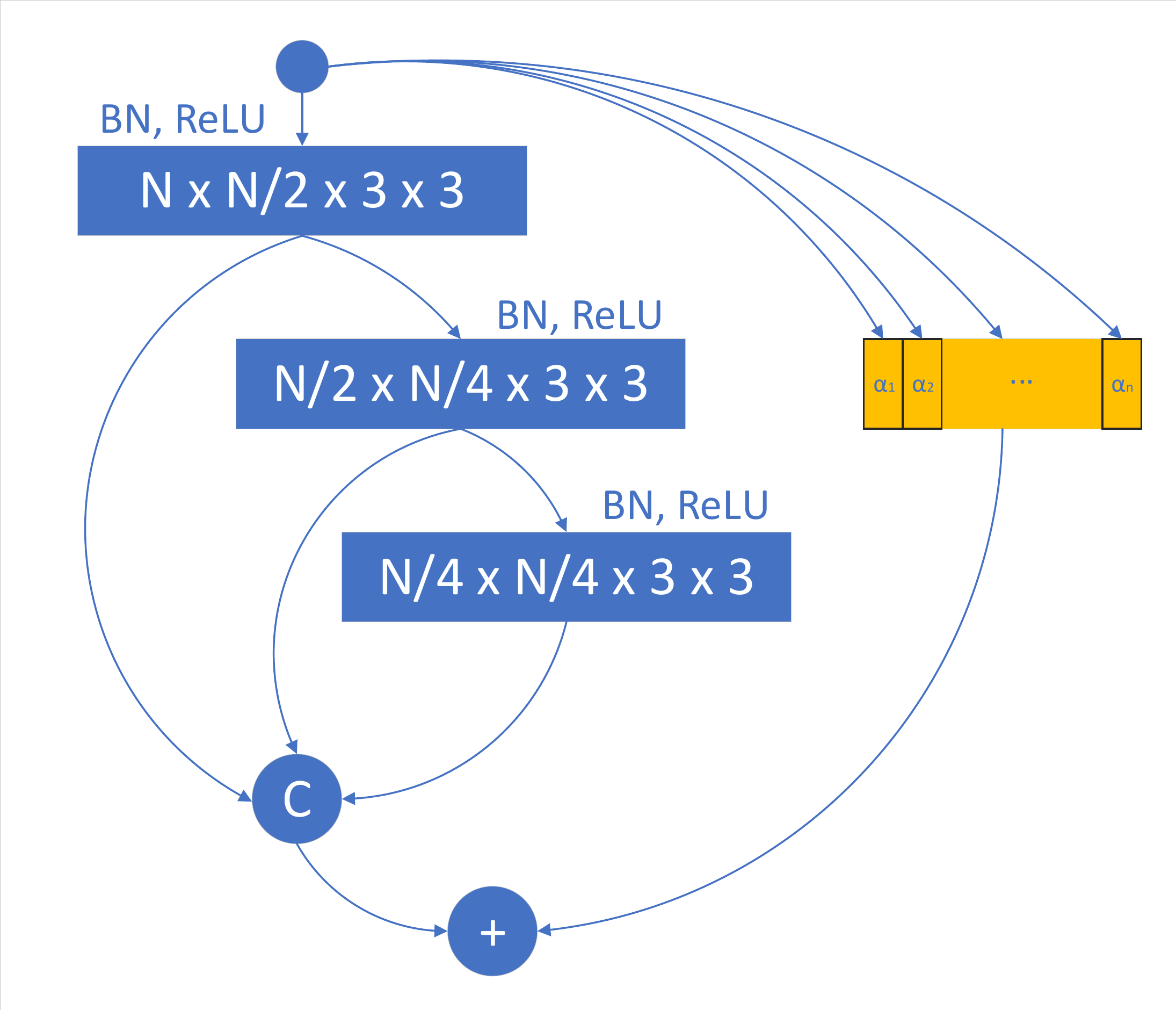}
\caption{The proposed gated skip connections with \textit{per-channel} learnable parameters for controlling the data flow for each channel within the module ($N$ is the number of input channels to each conv. layer).}
\label{fig:residul-block-alpha}
\end{figure}

Yet, from a practical perspective, many applications cannot fully enjoy the high accuracy demonstrated by recent advances. The reason is for this is twofold: (a) the bulk of current work assumes the abundance of computational resources (e.g. GPUs, memory, power) to run these models which for many applications are not available. (b) In many application domains (e.g. autonomous driving) accuracy is absolutely essential, and there is very little room for accuracy drop when, for example, more lightweight, compact, and memory efficient methods are used. 

Hence, although there are more and more methods proposed recently that achieve top performance in difficult benchmarks like the MPII \cite{andriluka14cvpr} and LSP \cite{Johnson10} and COCO~\cite{lin2014microsoft}, we argue that there is a significant gap literature as there are no methods which can get any close to this accuracy when there are memory (in terms of $\#$ of parameters) and computational power (in terms of flops) constraints. The focus of this work is to offer an improvement over the state-of-the-art under this setting.

Besides being challenging, the problem of human pose estimation under low memory and computing capacity has received little attention from the research community so far. 
To our knowledge there are only two very recent papers that make an attempt towards this direction: the works of \cite{bulat2017binarized} and \cite{tang2018quantized}.
Both methods aim at improving human pose estimation accuracy by introducing novel architectural changes at block~\cite{bulat2017binarized} and network~\cite{tang2018quantized} level. Our work improves upon these two works by looking into a component of the network architecture that is used ``as is'' in most of recent works: skip connections. While residual connections proved to be quintessential for achieving high accuracy within Fully Convolutional Networks, in this work we re-analyze this design choice in the context of human pose estimation and show that with simple improvements significant gains can be achieved both in terms of accuracy and efficiency for the whole complexity (memory, and flops) spectrum.

In particular, we make the following contributions: 
\begin{itemize}
    \item 
    We propose gated skip connections with per-channel learnable parameters to control the data flow for each channel within the module. This has the simple effect to learn how much information from the previous stage is propagated into the next one \textit{per channel} and encourages each module learn more complicated functions.  
    \item
    We introduce a hybrid network structure that combines the HourGlass \cite{newell2016stacked} and U-Net \cite{ronneberger2015u} architectures. The newly proposed architecture minimizes the number of identity connections within the network, and is shown to increase the performance within the same number of parameters budget.
    \item
    We report state-of-the-art results across the whole spectrum of $\#$ parameters and FLOPS. Our method is capable of producing a reduction of 65\% in model size and complexity (i.e. more twice as fast) with no decrease in performance when compared to the original HourGlass network. A larger version of our model achieves state-of-the-art results on the MPII and LSP datasets.
\end{itemize}

\section{Related work}\label{sec:realted-work}
Here we review the related work, first for efficient neural networks, where we define efficiency in terms of number of FLOPs~\ref{ssec:efficient-cnns} and then for human pose estimation~\ref{ssec:human-pose-estimation}.

\subsection{Efficient neural networks}\label{ssec:efficient-cnns}

It is now widely accepted that the advent of more powerful computational resources, and availability of large amount of annotated data are at the root of the recent successes of deep learning.
However, aside from these, it is architectural improvements that have made it possible to reach the remarkable levels of performance attained on a variety of tasks, ranging from image classification~\cite{he2016deep,he2016identity} to fine-grained recognition~\cite{newell2016stacked,bulat2016human,wei2016convolutional}.

In particular, the depth of deep networks is a crucial aspect of their performance. Training these has been possible since AlexNet~\cite{krizhevsky2012imagenet}, and then VGG~\cite{simonyan2014very}, the success of which has led to deeper neural networks.
However, this ever increasing depth also makes for harder to train networks.
This can be explained by several factors: i) increase in the number of parameters, ii) gradient vanishing or exploding.
However, recent architectural changes made it possible to train very deep networks.
The most important of these change was the introduction of skip connections within neural networks, which allow information to flow more easily within the network, both at forward time (activation) or during the backward pass (during which gradient can flow more easily, thus alleviating vanishing or exploding gradients phenomena).

ResNet~\cite{he2016deep} uses these within blocks of convolution --with or without bottleneck--. 
Finally, pushing this to the extreme, more recently, DenseNet~\cite{huang2016densely} proposes to introduce one to all connections  between a convolutional block and its successors within the same block.

While the above methods focus on performance (typically classification accuracy), neural networks can also be made more efficient in terms of computation. For instance, in \cite{iandola2016squeezenet}, the authors propose to leverage \(1\times1\) convolutions and skip connections to achieve performance similar to AlexNet on ImageNet but with a fraction of the parameters. MobileNet~\cite{howard2017mobilenets} builds on the same principle but explicitly parametrizes the convolutions as separable ones (e.g. their kernels can be expressed as the sum of rank one tensors).

The same concept has been instrumental in improving the state-of-the-art for human pose estimation, e.g. by introducing skip connections between the encoder and decoder parts of U-Nets. We detail these changes in the following subsection.

\begin{figure*}[t]
\centering
\begin{subfigure}[t]{\textwidth}
\centering
\includegraphics[width=16.0cm]{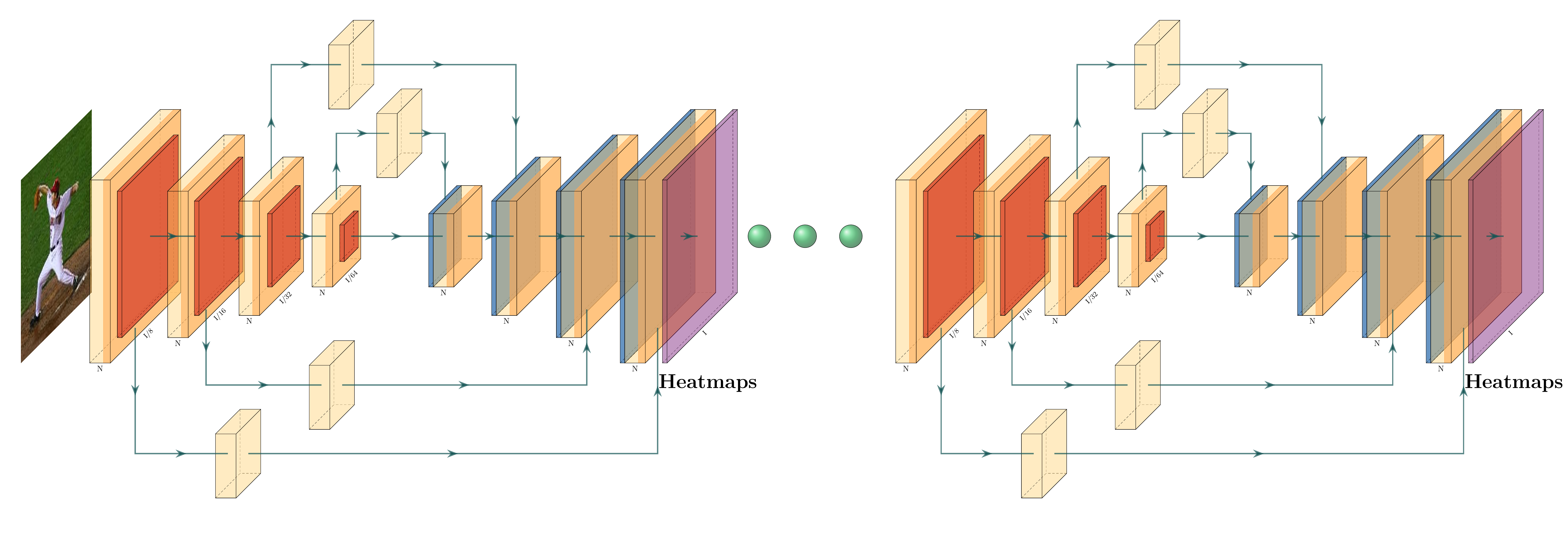}
\end{subfigure}\\
\begin{subfigure}[t]{.3\textwidth}
\includegraphics[width=4.0cm]{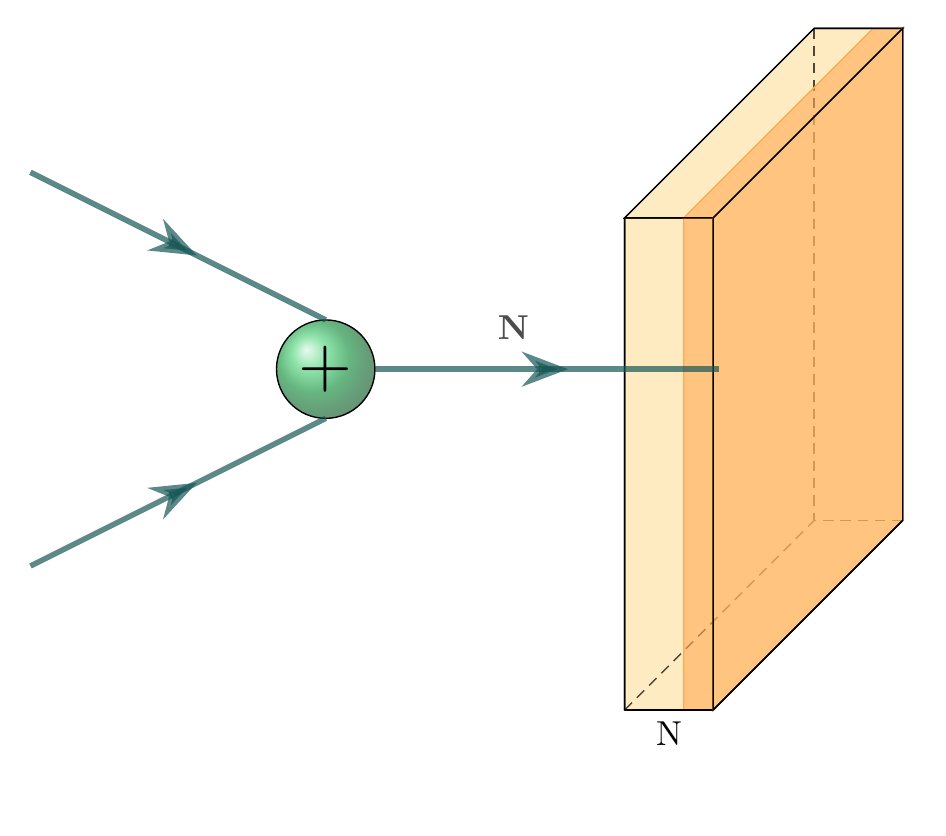}
\caption{\textbf{Baseline~\cite{newell2016stacked}}. The features coming from the encoder are merged in the decoder using element-wise summation, resulting in the same dimensionality \(N\).}
\label{sfig:block1}
\end{subfigure}
\hfill
\begin{subfigure}[t]{.3\textwidth}
\includegraphics[width=4.0cm]{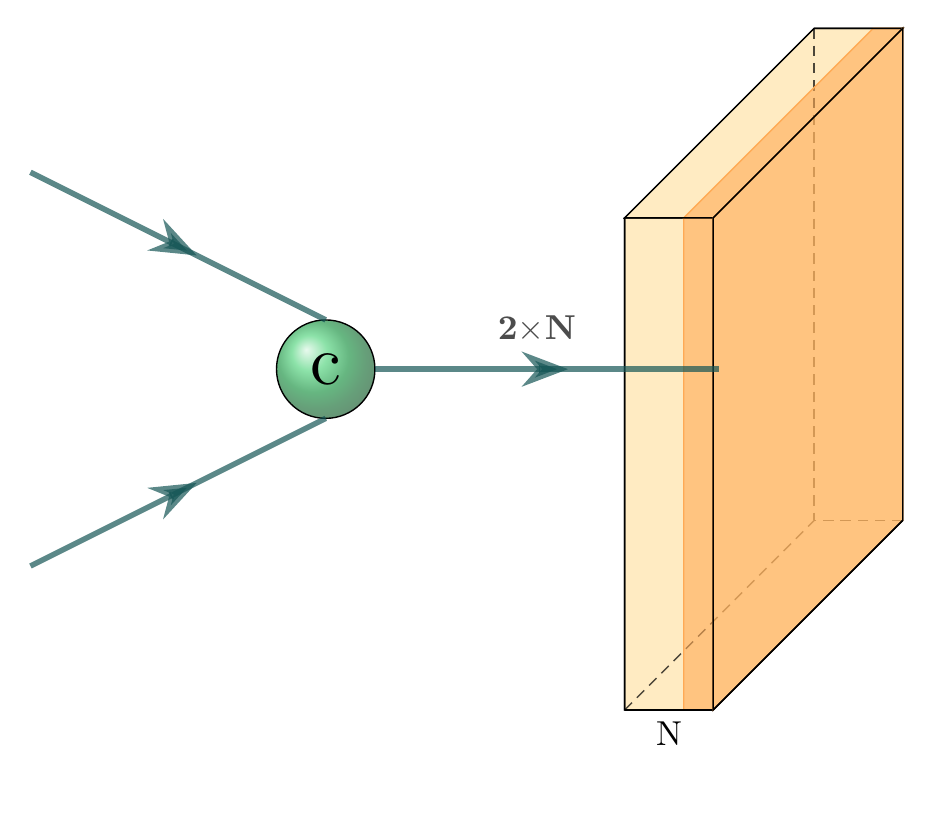}
\caption{\textbf{Proposed}. The features are first concatenated; a convolutional layer with a $3\times3$ kernel then reduces their dimensionality back to $N$.}
\label{sfig:block2}
\end{subfigure}
\hfill
\begin{subfigure}[t]{.3\textwidth}
\includegraphics[width=4.0cm]{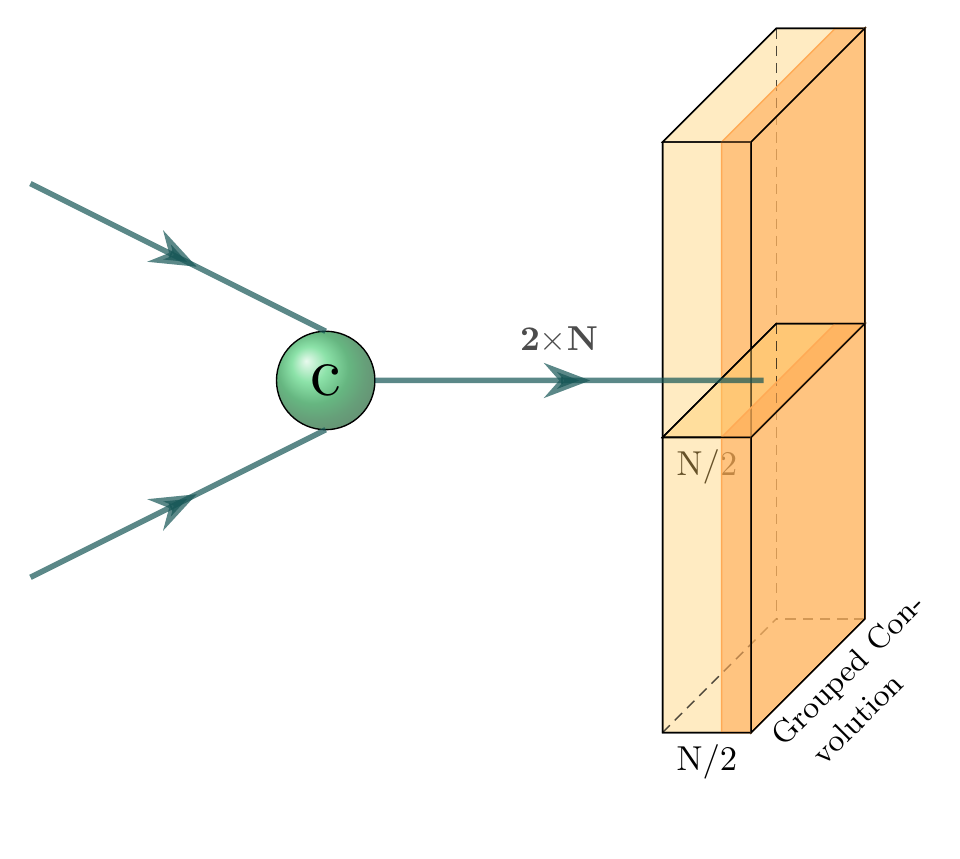}
\caption{\textbf{Proposed}. The features are concatenated and then processed using a grouped convolutional layer with a kernel of size $3\times3$. }
\label{sfig:block3}
\end{subfigure}
\caption{\textbf{Overall network architecture.} We depict the overall network architecture (top-row) and detail the various ways in which features coming from the skip connections are agregated i) in existing work~(\ref{sfig:block1}), (ii) our proposed concatenation approach~(\ref{sfig:block2}) and (iii) our proposed concatenation with grouped convolution~(\ref{sfig:block3}). Each yellow rectangular cuboid depicts the hierarchical residual module shown in Fig.~\ref{fig:residul-block-alpha}, the red one a max-pooling layer and the blue one a nearest neighbour upsampling operation. For our method the number of parameters is varied by changing the width (i.e number of channels) and the number of stacks.}
\label{fig:overall-architecture}
\vspace{-0.5cm}
\end{figure*}

\subsection{Human pose estimation}\label{ssec:human-pose-estimation}

Current state-of-the-art on single person pose estimation is held by variants of the so-called HourGlass~\cite{newell2016stacked,wei2016convolutional,bulat2016human,ke2018multi,chu2017multi,yang2017learning,chen2017adversarial} and U-Net architectures~\cite{ronneberger2015u,tang2018deeply,tang2018cu}. Both HourGlass and U-Net architectures consist of a stack of encoder-decoder Fully Convolutional Networks (see Fig.~\ref{fig:overall-architecture}) with skip connections between the encoder and the decoder part. On each skip connection between them a residual block is usually placed. The resolution is decreased, and respectively increased, 4 times (from $64\times64$px to $4\times4$px. 

In~\cite{tompson2015efficient}, the authors propose a coarse-to-fine learning mechanism where an initial coarse prediction is refined at a later stage by~\textit{zooming-in} into a region of interest and predicting a correction (expressed as an offset from the coarse detection). The work of Lifshitz et al.~\cite{lifshitz2016human} proposes to localize the landmarks using a dense voting technique where joint probabilities are learned from relative keypoint locations. In~\cite{belagiannis2017recurrent}, the authors introduce a hybrid architecture that combines a normal feed-forward model with a recurrent block. In a similar spirit with~\cite{newell2016stacked}, Wei et al.~\cite{wei2016convolutional} propose a 6-stack neural network to detect and gradually refine the keypoint predictions. In~\cite{bulat2016human}, the authors attempt to improve the localization process by diving the keypoint detection task into two sup-problems: detection and regression. At the first stage they detect only the visible landmarks using a part detection network, while at the second one they regress jointly the position of all keypoints, both visible and occluded. 

More recent methods attempt to combine the HG based architecture with attention mechanisms~\cite{chu2017multi}, feature pyramids~\cite{yang2017learning} and adversarial training~\cite{chou2017self,chen2017adversarial}. The work of~\cite{chu2017multi} uses a Conditional Random Field in order to model the correlations among nearby regions combining a holistic attention model with a body part one. This way the network learns to focus on both global and local details. In order to enforce a stronger model, in~\cite{chou2017self}, the authors propose an adversarial training approach where the pose estimator has the role of a generator. At training time a discriminator is used to assess the quality of the produced heatmaps. A similar approach is followed in~\cite{chen2017adversarial}, where a discriminator is used to discern between feasible and biologically unfeasible poses. Yang et al.~\cite{yang2017learning} proposes a Pyramid Residual Module to improve the scale invariance of the models. The Pyramid Residual Module learns a series of convolutional filters at various input scales, on features obtained using different sub-sampling ratios. In~\cite{ke2018multi}, the authors propose a series of architectural enhancements aimed at improving the overall network robustness such as the addition of a multi-scale supervision, a structure-aware loss and a keypoint masking technique aimed at increasing the accuracy of the occluded points. With the goal of strengthening the intrinsic human body model learned by the network,~\cite{tang2018deeply} introduces a compositional model that learns the relation between various body parts.

While most of the prior works focus on increasing the network accuracy without enforcing any strict computational requirements, herein we attempt to obtain a neural network that performs well across different level of computational resources. To our knowledge, the only papers that have similar aims with our work are~\cite{bulat2017binarized}  and~\cite{tang2018cu}.
\cite{bulat2017binarized} proposes to improve the speed of human pose estimation by fully binarizing the features and the weights of a given network. The work of~\cite{tang2018cu} combines dense connections with the HG model improving the overall accuracy and speed. 

Our work is different to both~\cite{bulat2017binarized}  and~\cite{tang2018cu} (as well to all aforementioned papers on human pose estimation) because it improves upon skip connections, one of the most fundamental component of deep architectures which is used ``as is'' by all methods for human pose estimation. We show that the proposed enhancements significantly improves the overall network performance within the same computational budget.

\section{Method}\label{sec:method}
Here we introduce in detail our method (Section~\ref{ssec:revisiting} and~\ref{ssec:network-architecture}), the network architecture used as well as the implementation and training details (Section~\ref{ssec:training}).
\begin{figure*}[!htbp]
    \begin{subfigure}[t]{0.23\textwidth}
    \centering
    \includegraphics[width=1\linewidth]{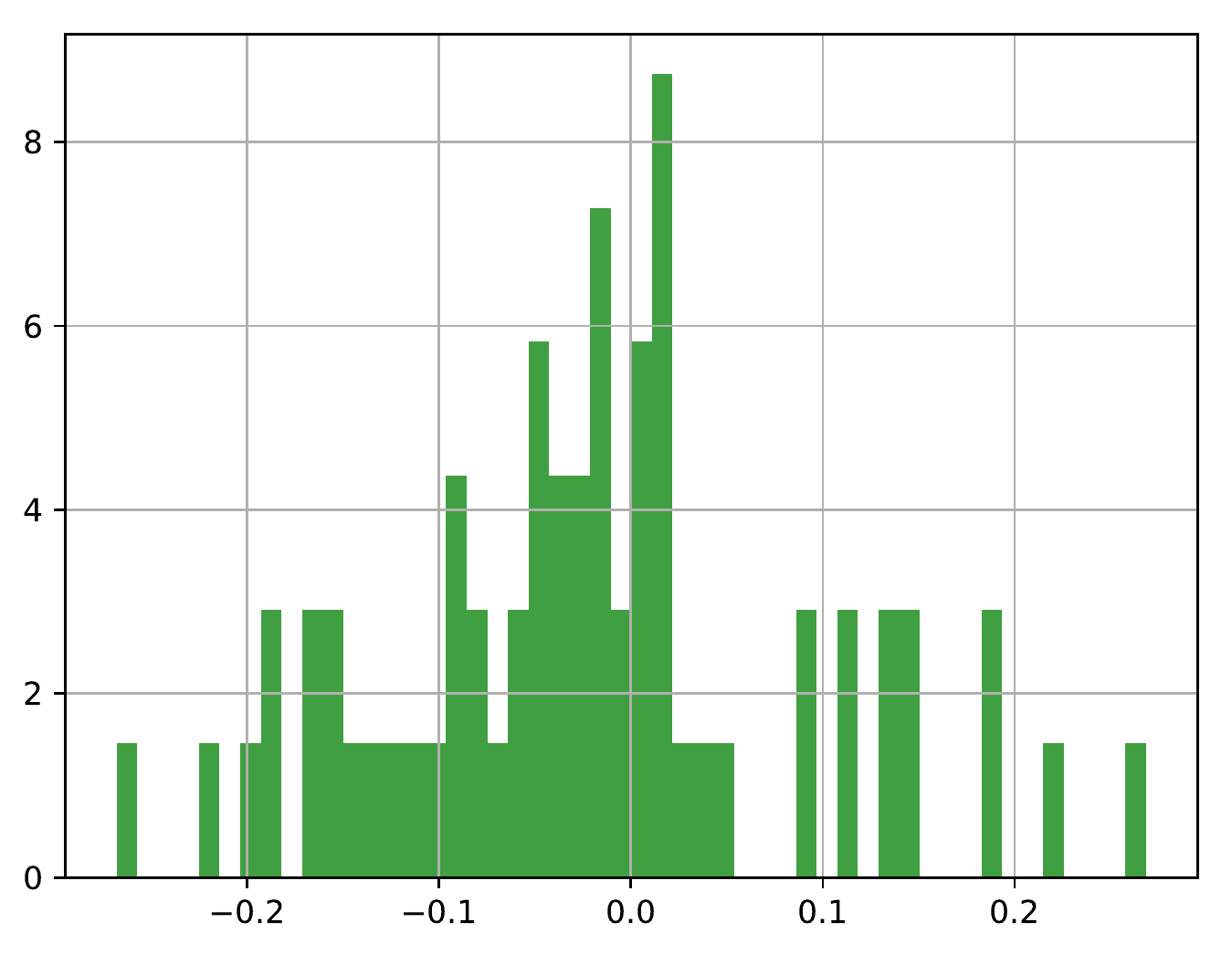}
    \end{subfigure}
    ~
    \begin{subfigure}[t]{0.23\textwidth}
    \centering
    \includegraphics[width=1\linewidth]{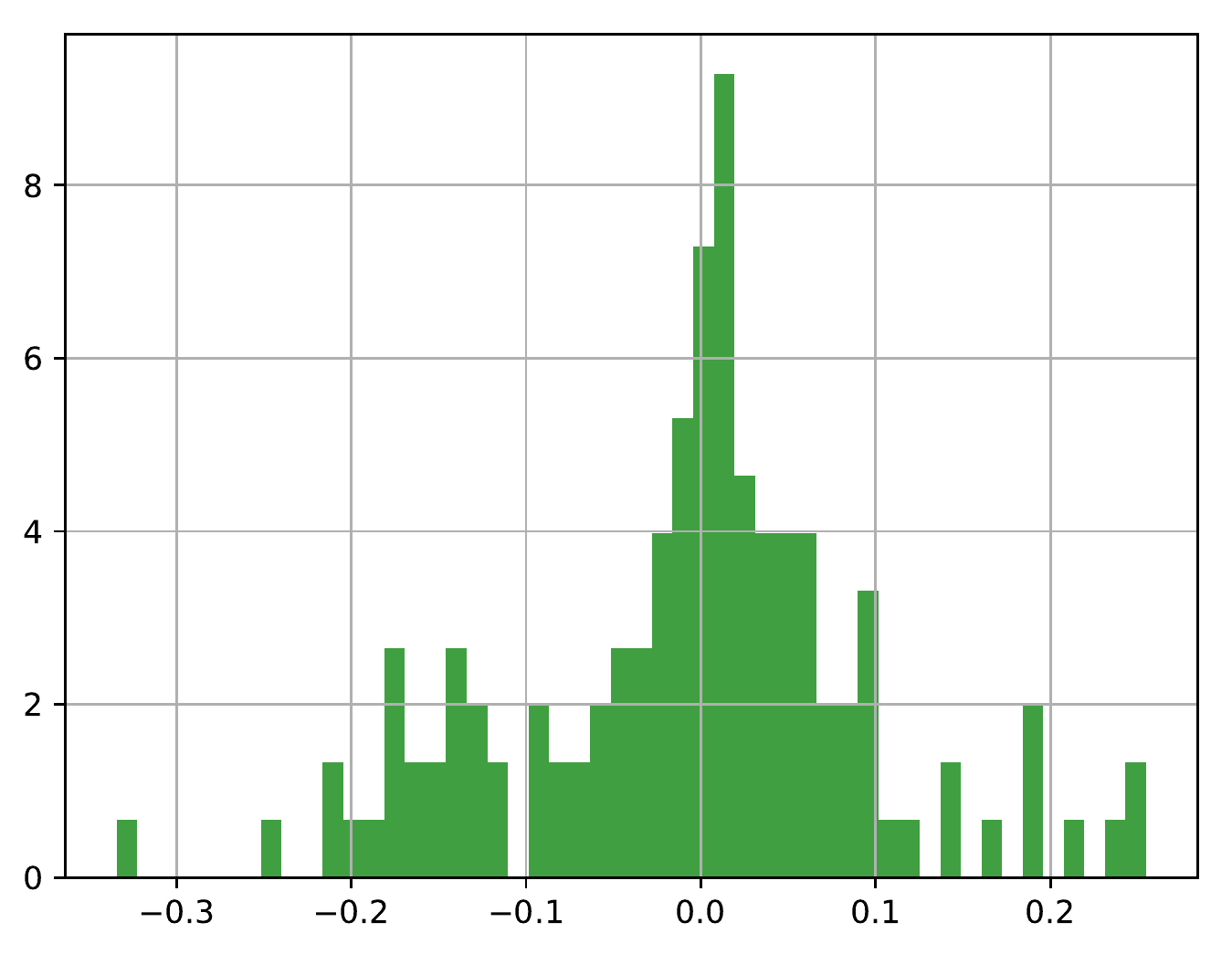}
    \end{subfigure}
    ~
    \begin{subfigure}[t]{0.23\textwidth}
    \centering
    \includegraphics[width=1\linewidth]{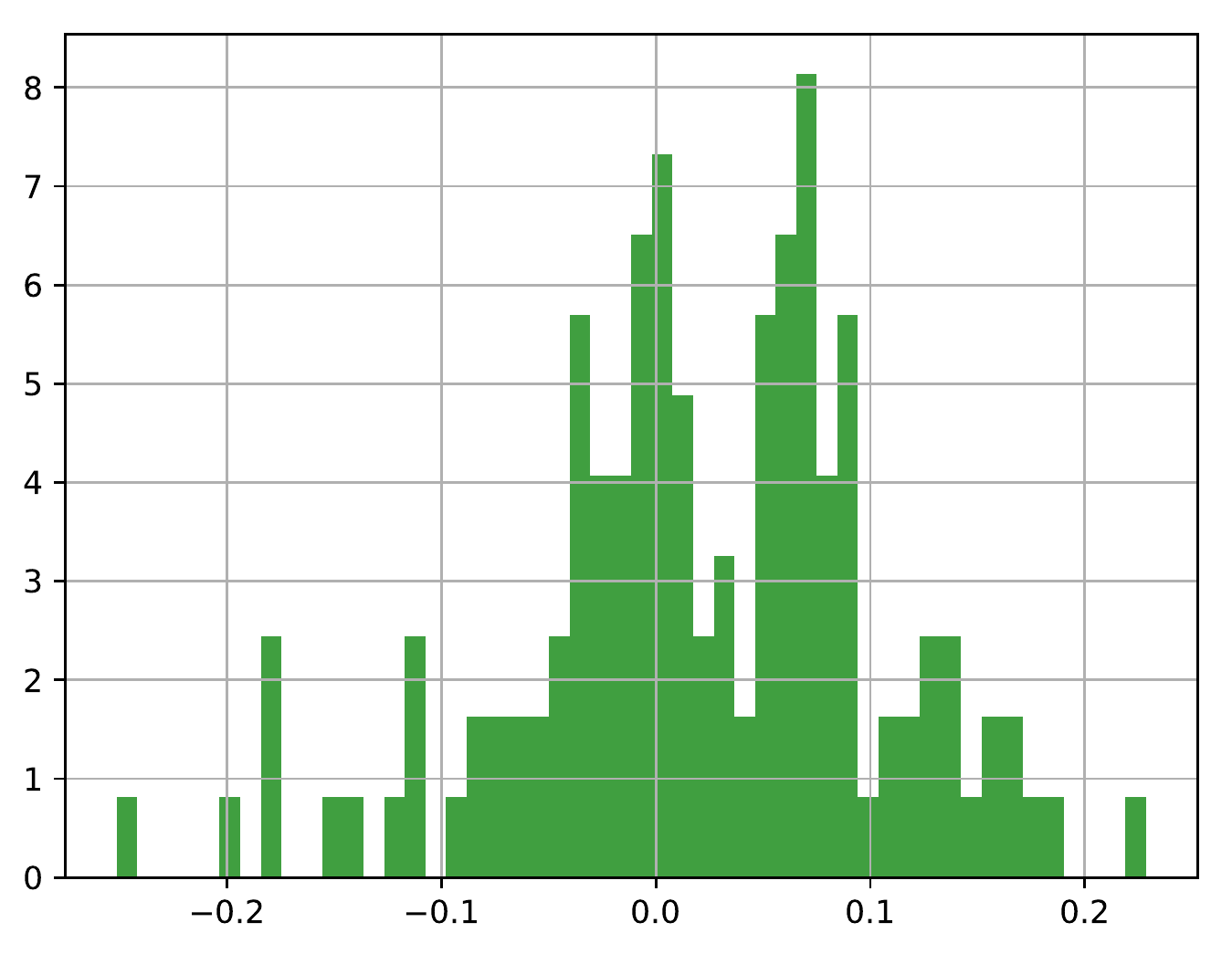}
    \end{subfigure}
    ~
    \begin{subfigure}[t]{0.23\textwidth}
    \centering
    \includegraphics[width=1\linewidth]{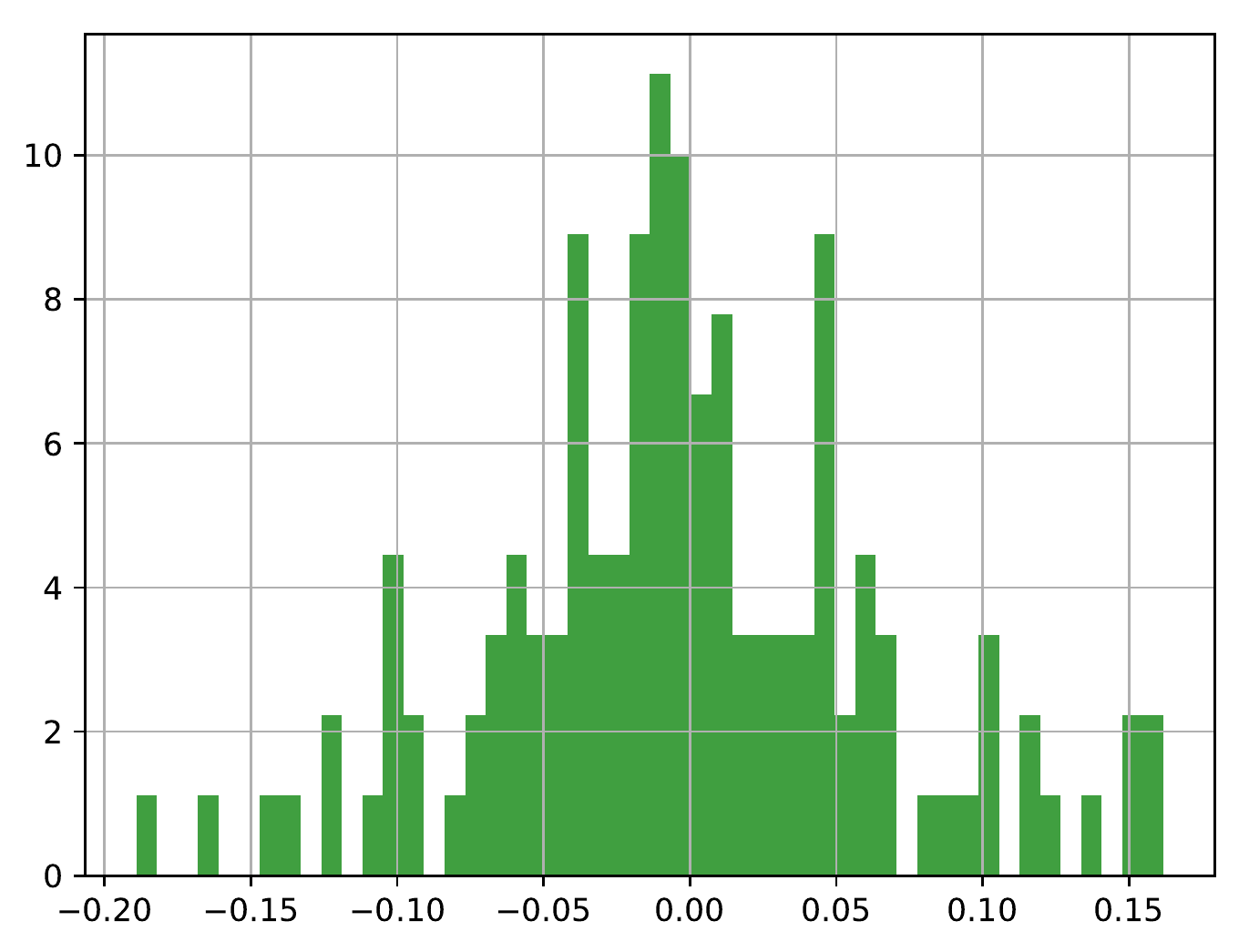}
    \end{subfigure}

    \caption{Output distribution of the scaling factor $\alpha\in\mathbb{R}^{N}$ for various blocks through the network from the bottom to top (left to right). Notice that most of the values are clustered around $0$ and contained in the interval $[-0.1, 0.1]$.}
    \label{fig:alpha-distribution}
\end{figure*}

\begin{figure}[!htbp]
    \centering
    \begin{subfigure}[t]{.23\textwidth}
    \centering
    \includegraphics[width=\linewidth]{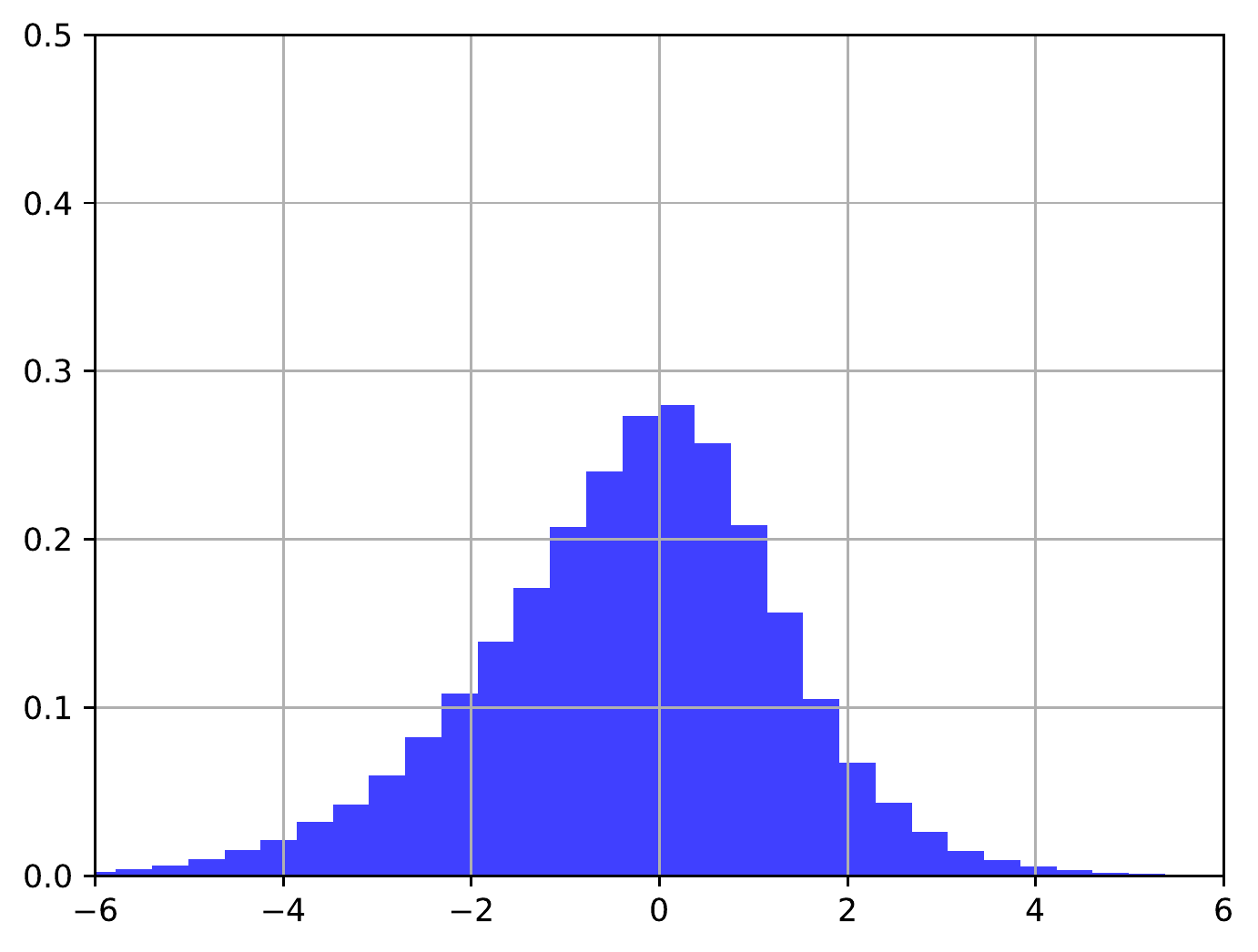}
    \caption{Features distribution at the output of the \textbf{baseline} residual block (i.e. the skip connection is implemented using the Identity function), immediately after summation.}
    \label{fig:weights-distribution-sum-old}
    \end{subfigure}
    ~
    \begin{subfigure}[t]{.23\textwidth}
    \centering
    \includegraphics[width=\linewidth]{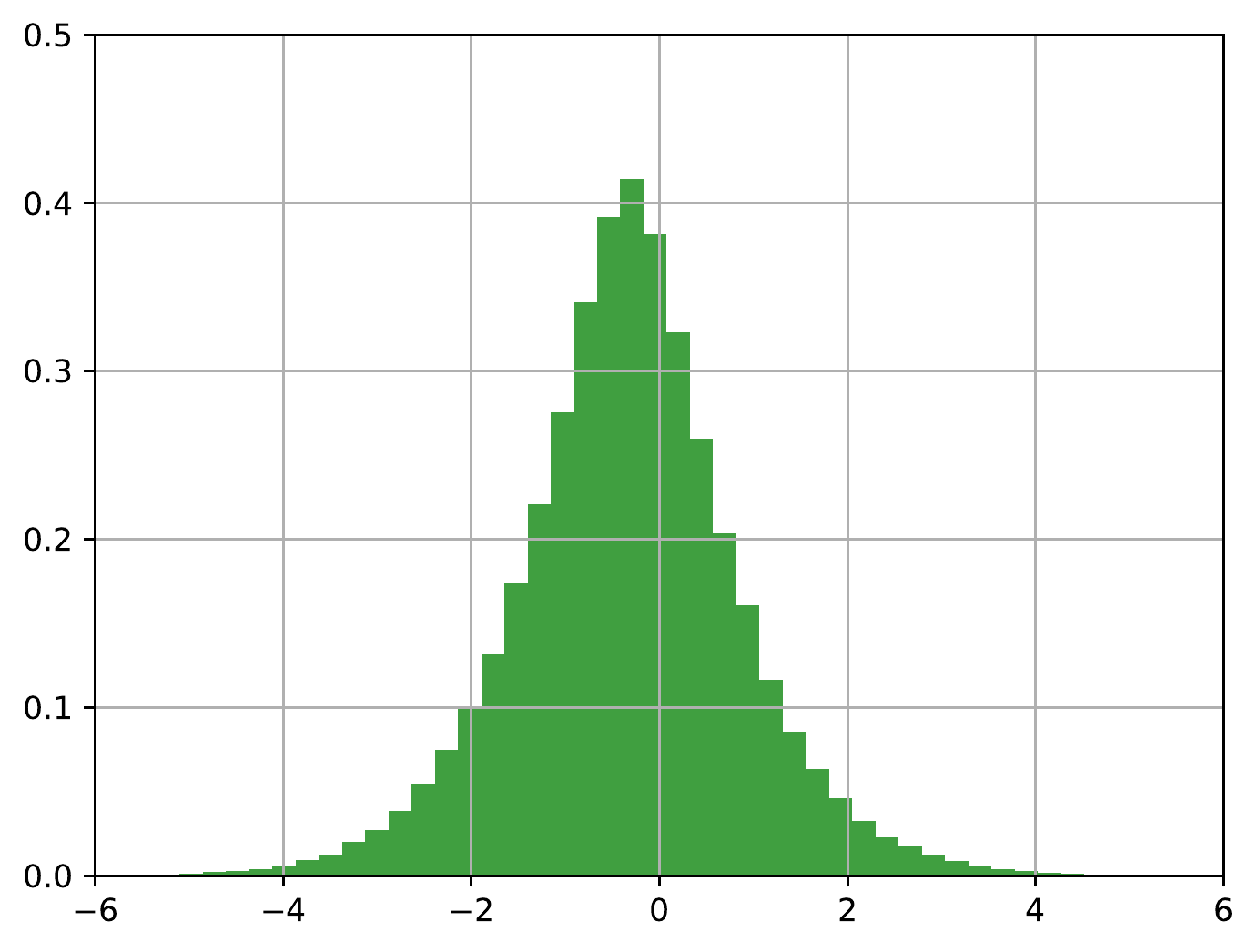}
    \caption{Features distribution at the output of the residual block that uses the \textbf{proposed} soft gating function (see Fig.~\ref{fig:residul-block-alpha}), immediately after summation.}
    \label{fig:weights-distribution-sum-new}
    \end{subfigure}
    \caption{\textbf{Comparison of the distribution of the features at the output of the residual block} for the baseline (Fig.~\ref{fig:weights-distribution-sum-old}) and the proposed approach (Fig.~\ref{fig:weights-distribution-sum-new}). Notice that the proposed method, with the help of soft gating function, can preserve the function learned by the residual module $l$. In contrast, the baseline module is forced to incorporate all of the information coming from the previous module, limiting as a consequence its representational power.}
    \label{fig:weights-distribution-after-sum}
\end{figure}

\begin{figure}[!htbp]
    \centering
    \begin{subfigure}[t]{.48\textwidth}
    \centering
    \includegraphics[width=.475\linewidth]{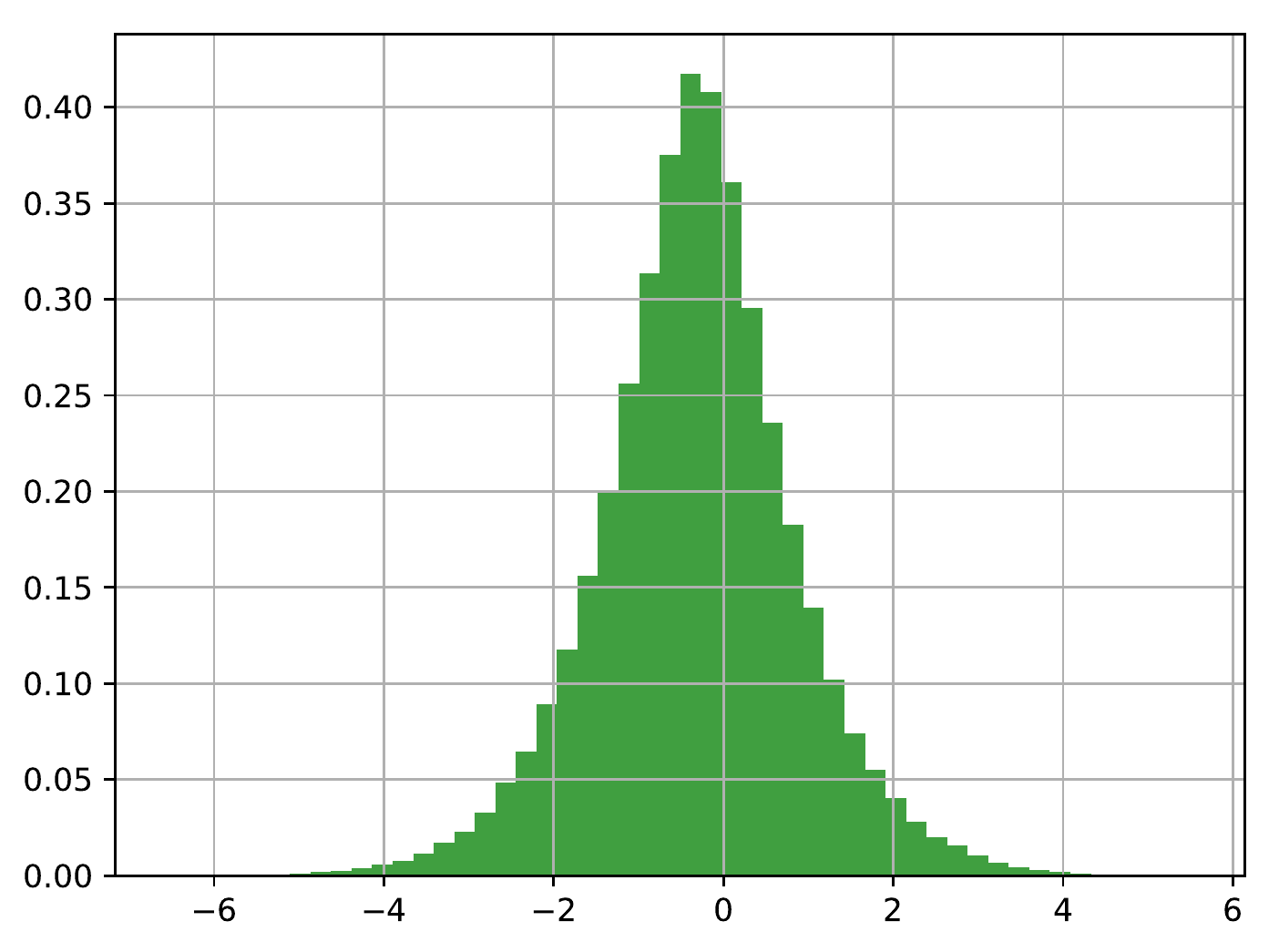}
    \hfill
    \includegraphics[width=.475\linewidth]{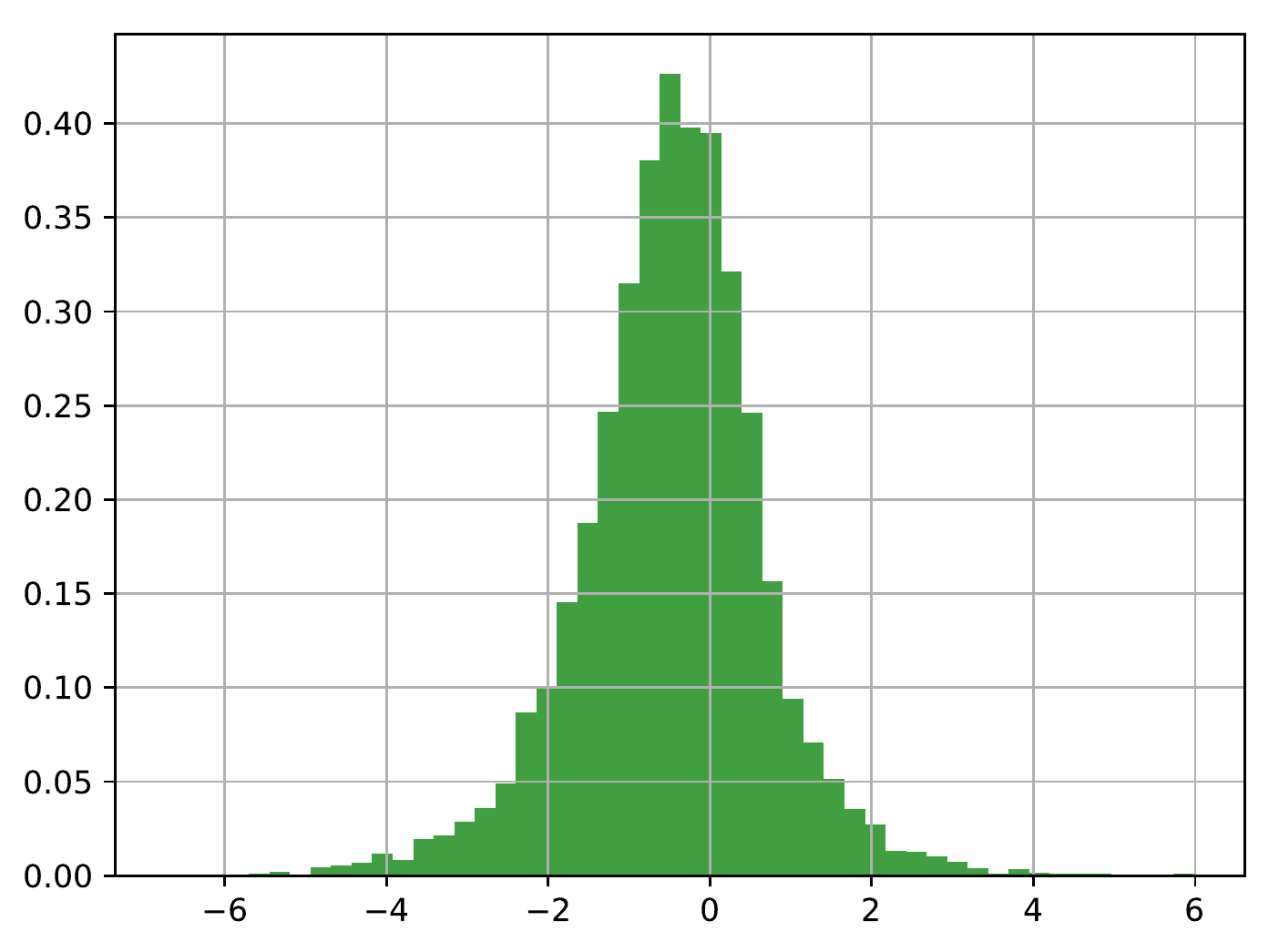}
    \caption{Features distribution after the concatenation stage inside the residual block.}
    \label{fig:weights-distribution-1}
    \end{subfigure}
    \vskip\baselineskip
    \begin{subfigure}[t]{.48\textwidth}
    \centering
    \includegraphics[width=.475\linewidth]{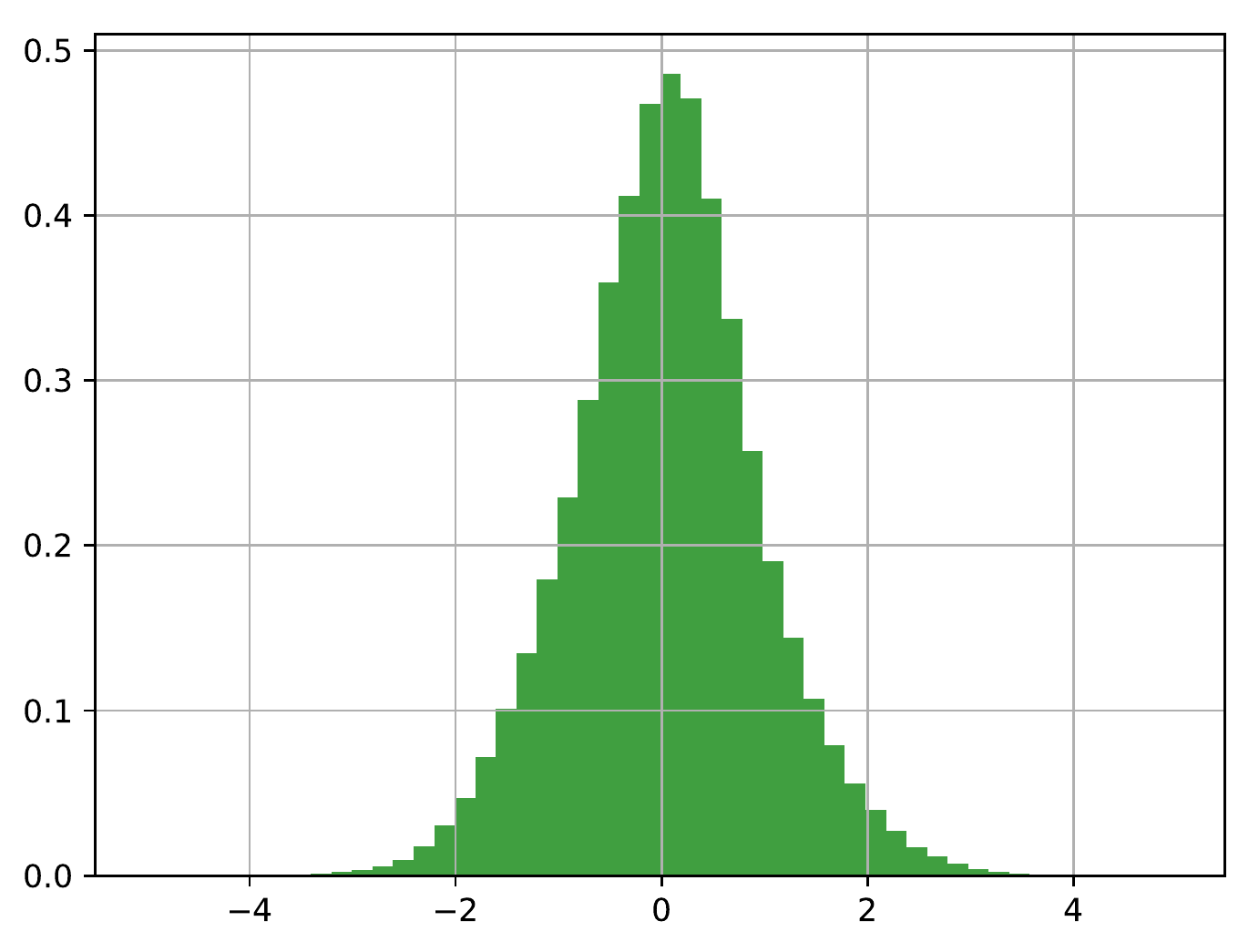}
    \hfill
    \includegraphics[width=.475\linewidth]{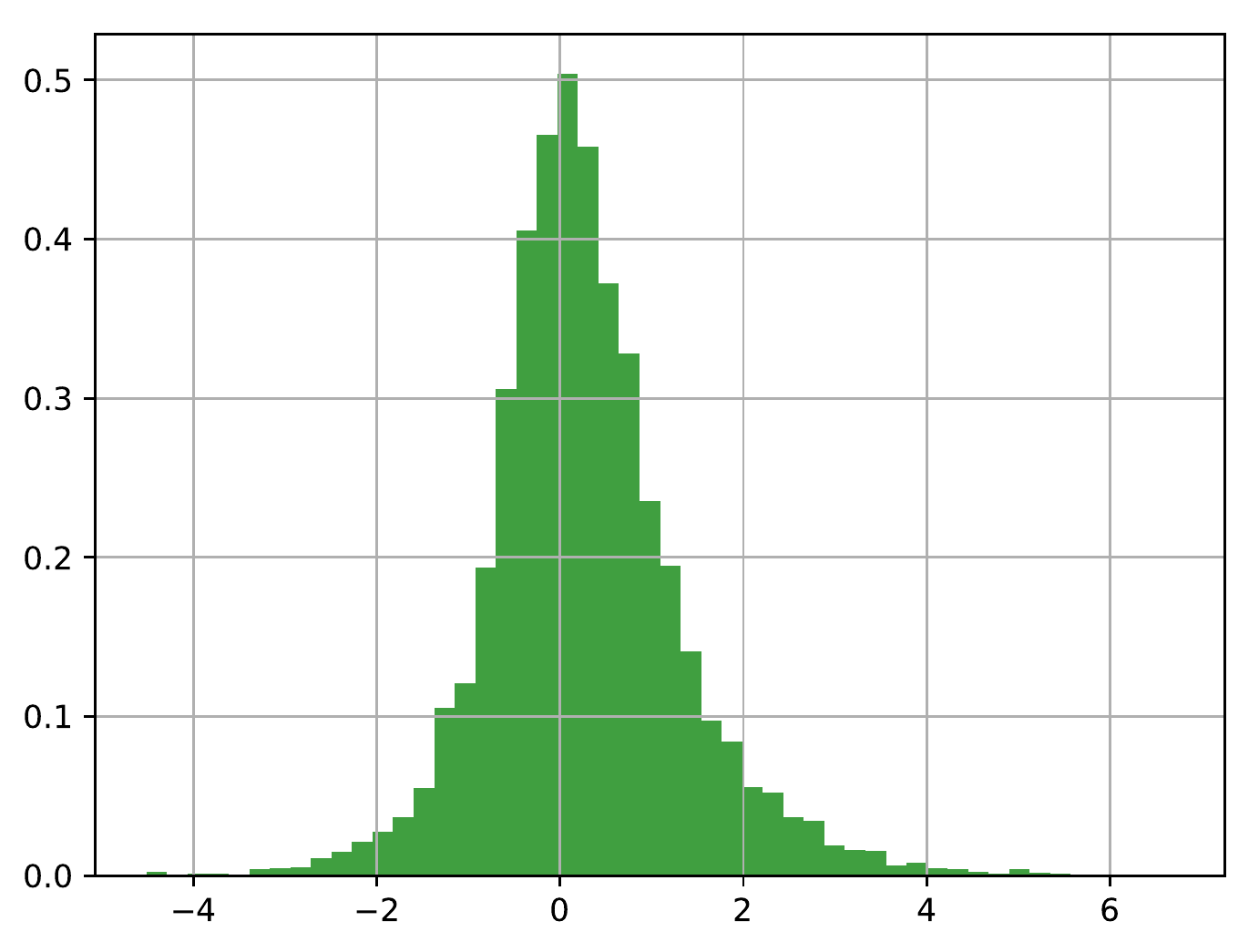}
    \caption{Features distribution on the skip connection before scaling them using $\alpha$.}
    \label{fig:weights-distribution-2}
    \end{subfigure}
    \vskip\baselineskip
    \begin{subfigure}[t]{.48\textwidth}
    \centering
    \includegraphics[width=.475\linewidth]{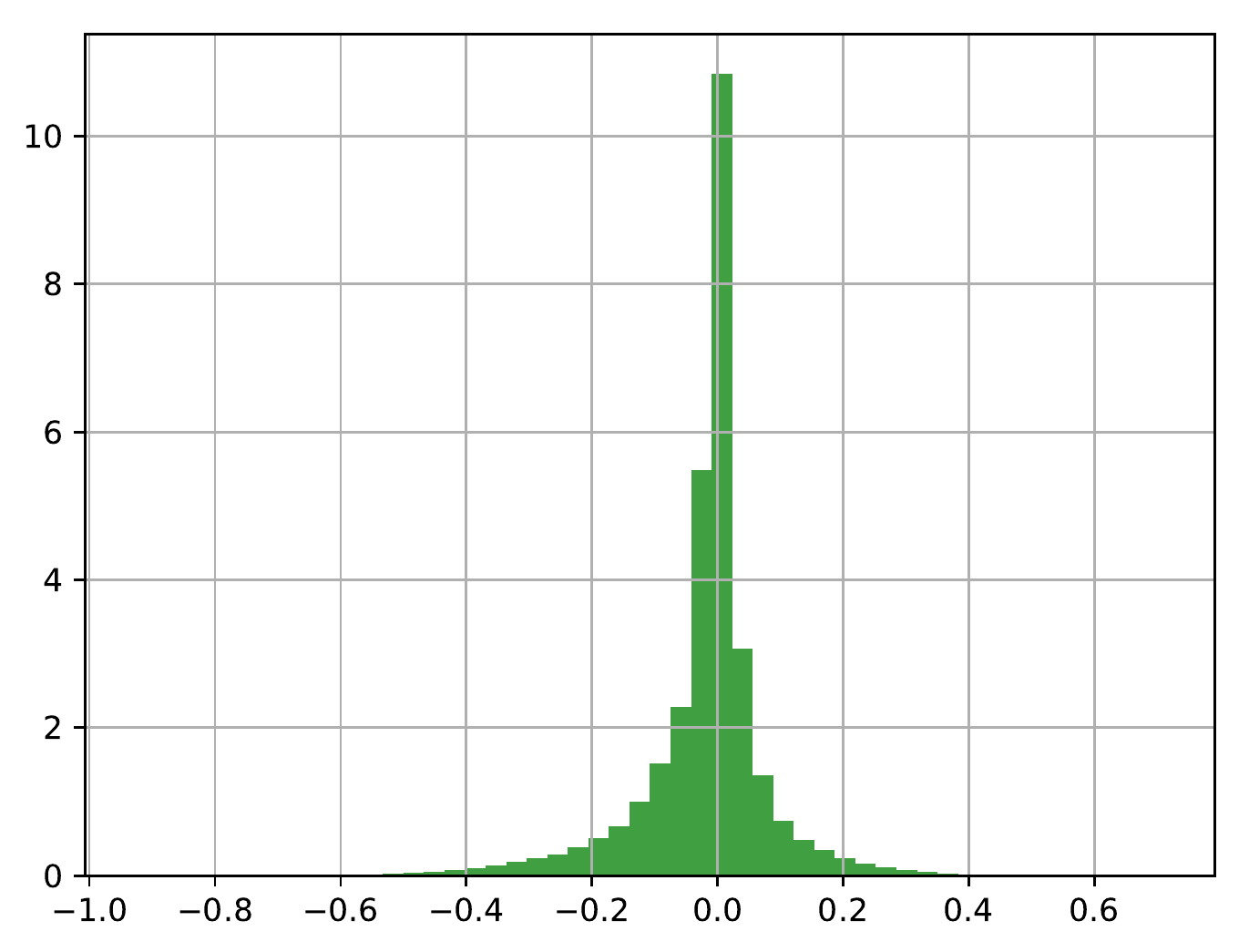}
    \hfill
    \includegraphics[width=.475\linewidth]{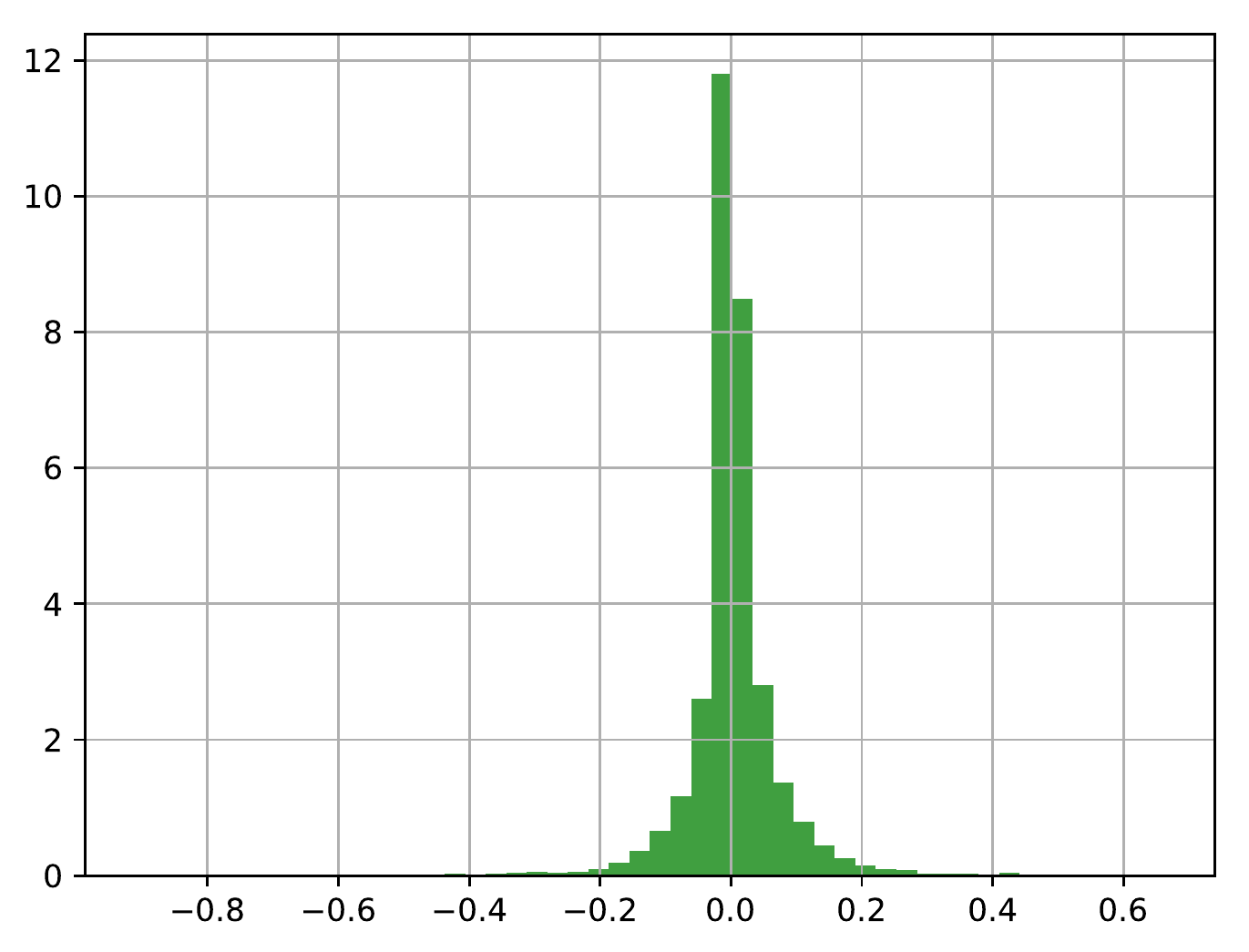}
    \caption{Features distribution on the skip connection after scaling them using~$\alpha$.}
    \label{fig:weights-distribution-3}
    \end{subfigure}
    \caption{\textbf{Comparison of the distribution of the features} in various scenarios for two layers from the bottom(left column) and top of the network(right column). Notice that after applying the scaling factors $\alpha$ to the features from the skip connection (second row) most of the values are scaled toward $0$ (third row). This suggests that only a handful of information (i.e. ``residual information'') is actually useful at the next stage. Combining directly the features coming from the module and the skip connection may have undesirable effect and hinder the overall performance since the function will have to learn a more incremental step due to the avalanche of information from the step $t-1$.}
    \label{fig:weights-distribution}
\end{figure}

\subsection{Soft-gated residual connections}~\label{ssec:revisiting}
As detailed in Section~\ref{ssec:efficient-cnns}, one of the key aspects that made training very deep neural networks possible was the introduction of residual networks~\cite{he2016deep}. Residual connections have since become a ubiquitous part of current state-of-the-art neural network architectures and are often considered a quintessential aspect that drives their accuracy. Despite this, we argue here that, at least for some cases, the presence of an identity connection may have undesirable effects and hinder the performance of the model.

In the quest for training very deep neural networks, the work of~\cite{he2016identity,srivastava2015highway} explores the effect of using a \textit{hard gating} function $g(\myvector{x}) = \sigma (\mytensor{W}_g\myvector{x} + \myvector{b}_g)$, where $\sigma(\myvector{x}) = \frac{1}{1-e^{-x}}$ is the sigmoid function, $\mytensor{W}_g$ and $\myvector{b}_g$ the weights and respectively the bias of a given \textit{hard gating} transformations. Typically $g(x)$ is implemented using a convolutional layer with a kernel size of $1\times1$. When tested on ResNet architectures, these changes lead to sub-par results or even fail to converge~\cite{he2016identity}.

Currently, despite their relatively shallow nature at a stack level, all HourGlass-like architectures take the benefit of using skip connections for granted. 
Herein, we argue that this is not universally true and explore their effect in the context of human pose estimation, showing that our method can outperform across various computational budgets, architectures that make extensive use of skip connections~\cite{newell2016stacked,tang2018cu} (see Section~\ref{sec:evaluation}).

It is already widely accepted that ResNets learn an unrolled iterative estimation~\cite{greff2016highway} with each residual unit learning a small correction with respect to the previous unit. As such, removing an arbitrary residual unit inside a macro-module leads to an insignificant drop in performance~\cite{jastrzkebski2017residual} as long as the first and the last units are kept. Since in an HourGlass stack, typically, at each resolution level inside the encoder and decoder a very small number of residual blocks are used (usually one), forcing the blocks to learn a correction with respect to the input hinders the learning process and goes against the finding from~\cite{jastrzkebski2017residual} that suggest that at the transition level between resolutions novel functions need to be learned by the network. By addressing this, we will show bellow that a 4 stacks HG network matches and outperforms all of the previous methods that normally use 8 or more stacks which suggests that our network can learn stronger and more diverse functions. 

In this work, we propose a novel way of improving the residual module using a \textit{channel-wise} soft gating mechanism defined as bellow:
\begin{equation}
    x_{l+1} = \myvector{\alpha} x_l + \mytensor{F}(x_l,\mytensor{W}_l),
\end{equation}
where $x_l \in \mathbb{R}^{C\times w \times h}$ are the input features from the previous layer, $\mytensor{W}_l$ is a set of weights associated with the $l$th residual block and $\mytensor{F}$ a residual function implemented using a set of convolutional layers (in this work using the module depicted in Fig.~\ref{fig:residul-block-alpha}). $\myvector{\alpha} \in \mathbb{R}^{C \times 1}$ is a channel-wise \textit{soft gate} (scaling factor) that is learned via backpropagation. 

\begin{table}[!htbp]
   \begin{center}
   \resizebox{1\linewidth}{!}{%
    \begin{tabular}{lcccccc}
    \toprule
    \textbf{Method} & \textbf{\# stacks} & \textbf{\# parameters} & \textbf{FLOPs $\times 10^{9}$} & \textbf{PCKh} \\
    \midrule
    CU-Net~\cite{tang2018cu} & 4 & 4.0M & 5.17 & 88.0\%  \\
    \textbf{Ours} & \textbf{4} & \textbf{3.4M} & \textbf{4.27} & \textbf{88.0}\%  \\
    \midrule
    CU-Net~\cite{tang2018cu} & 8 & 10.1M & 12.7 & 89.4\%  \\
    \textbf{Ours} & \textbf{4} & \textbf{8.5M} & \textbf{9.9} & \textbf{89.5}\%  \\
    \bottomrule
    \end{tabular}
    }
    \end{center}
    \caption{PCKh-based on the MPII validation set for the medium speed regime domain.}
    \label{tab:mpii_valid_normal}
\end{table}

Herein, we apply the newly proposed soft gating mechanism to the state-of-the-art module from~\cite{bulat2017binarized}, previously used for quantized neural networks. In the process we explore two different settings: (a) using a single soft gate for all channels and (b) learning a value for each input channel. As the results from Table~\ref{tab:mpii_valid_alpha} show, since different channels encode different types of information, the best results can be obtained using the channel-wise version that improves the overall performance against the baseline consisting of simple using the identity transformation (i.e. $\alpha=1$) by up to 1\%. 

To visualize the effect of the scaling factor we added in the skip connection to allow for soft-gating, we plot in Figure~\ref{fig:alpha-distribution} the output distribution of the scaling factors. Interestingly, we notice that the majority of the values are clustered around \(0\), which means that most of the information is not needed, or potentially even harmful for training. This phenomenon is observed across all layers of the network, regardless of the depth. These observations confirm the importance of this soft-gating parameter and its ability to filter redundant information. This is further reinforced in Fig.~\ref{fig:weights-distribution-after-sum} where the features learned by the proposed residual module variation are preserved after summation, since the scaling factors allows the module to select only the useful information from the previous stage. We also visualize how this scaling factors affects the distribution of the weights in the supplementary material. Notice that most of the features coming from the previous block are filtered by the introduced channel-wise scaling factor.

\begin{table}[!htbp]
   \begin{center}
    \begin{tabular}{lcc}
    \toprule
    \textbf{$\alpha$--skip}  & \textbf{learnable $\alpha$} &  \textbf{PCKh} \\
    \midrule
    \emph{baseline} &\xmark &  87.0\%  \\
    $\alpha=0.5$& \xmark&  87.4\%  \\
    $\alpha=0.1$ & \xmark&  87.5\%  \\
    $\alpha \in \mathbb{R}^{1}$ &\checkmark &  87.6\%  \\
    $\alpha \in \mathbb{R}^{N}$ &\checkmark &  \textbf{88.0}\%  \\
    \bottomrule
    \end{tabular}
    \end{center}
    \caption{PCKh-based comparison with state-of-the-art on the MPII validation set for different values and methods of computing the scaling factor $\alpha$.}
    \label{tab:mpii_valid_alpha}
\end{table}

\subsection{Improved network architecture}~\label{ssec:network-architecture}
Here we introduce a new hybrid network structure that combines the HourGlass~\cite{newell2016stacked} and U-Net~\cite{ronneberger2015u} architectures.
By minimizing the number of identity connections within the network, we are able to obtain superior performance with the same number of parameters as existing networks.

The HourGlass architecture as introduced in~\cite{newell2016stacked} and depicted in Fig.~\ref{fig:overall-architecture}, consists of a series of encoder-decoder macro-modules, where the predictions are gradually refined at each stage. Each residual module from a particular resolution level in the encoder is connected with its counterpart in the decoder. The connection can be realized either using an Identity function (U-Net) or using another residual module (HG). Typically in HG this data is fused using an element-wise summation. 

However, herein we argue that directly adding the features from two different distribution is suboptimal, as such we explore various way of aggregating the data coming from different sources (i.e. places in the network). As such we explore the following options: a) Concatenating the features and then processing them inside the residual module using a convolutional layer with $3\times3$ filters (Fig.~\ref{sfig:block2}), b) Concatenating the features and the combining them inside the residual module using a grouped convolutional layers, where the number of groups corresponds with the number of data sources, 2 in this case. Finally, we explore various block choices on the skip connection between the encoder and the decoder parts.

As the results from Table~\ref{tab:mpii_valid_connections_hg} show, for the same parameters budget (approx. 3.4M distributed across 2 stacks) concatenating the features and analyzing them jointly (groups=1) leads to the best results. improving on top of the baseline by 0.5\% when evaluated on the MPII validation set.

While we explored a series of different choices for the transform layer placed on the skip connections between the encoder and the decoder such as: [BatchNorm $\rightarrow$ ReLU $\rightarrow$ $1\times1$ Conv2D], [BatchNorm $\rightarrow$ ReLU $\rightarrow$ $3\times3$ Conv2D], [$1\times1$ Conv2D] etc we found no noticeable differences between them across multiple runs as long as the number of parameters across the entire network stayed roughly the same. This suggests that the layers found on the big-skip connections simple learn a feature projection.

\begin{table*}[!htbp]
	\begin{center}
   {\setlength{\tabcolsep}{12pt}
   \resizebox{0.9\linewidth}{!}{%
		\begin{tabular}{l|ccc|ccc}
			\toprule
			 & \multicolumn{3}{|c|}{\textbf{CU-Net~\cite{tang2018cu}}} & \multicolumn{3}{c}{\textbf{Ours}}\\
			\cline{2-7}
			\multirow{-2}{*}{\textbf{\# parameters}} & \textbf{\# stacks} & \textbf{FLOPs $\times 10^{9}$} & \textbf{PCKh} & \textbf{\# stacks} & \textbf{FLOPs $\times 10^{9}$} & \textbf{PCKh} \\
			\cline{1-7}
			  0.5M & 2 & 0.77 & \textbf{81.6}\% & 1 & 0.8  & \textbf{81.6}\%\\
			  1.0M & 2 & 1.68 & 84.2\% & 2  & 1.39 & \textbf{84.8}\%\\
			  1.4M & 2 & 2.14 & 85.6\% & 2 & 1.75& \textbf{85.7}\%\\
			  1.9M & 2 & 2.7 & 86.0\% & 2 & 2.43 & \textbf{86.4}\%\\
			  2.4M & 2 & 3.5 & 86.3\% & 2 & 3.1& \textbf{86.7}\%\\
			  2.9M & 2 & 4.15 & 86.6\% & 2 & 3.75 & \textbf{87.1}\%\\
			\bottomrule                                                                        
		\end{tabular}
	}}
	\end{center}
	\caption{Comparison against the state-of-the-art method of~\cite{tang2018cu} across the high speed regime domain. Notice that our method consistently outperforms~\cite{tang2018cu} up to $0.5$\% on the MPII validation set, while been less computationally demanding. The number of FLOPs for both networks is estimated using an input image $256\times256$px.}
	\label{tab:mpii_valid_comparison}
\end{table*}

\begin{table}
   \begin{center}
    \begin{tabular}{lc}
    \toprule
    Connection type  &  \textbf{PCKh} \\
    \midrule
    Baseline (Fig.~\ref{sfig:block1})  &  87.5\%  \\
    Proposed concat. (Fig.~\ref{sfig:block2})  &  88.0\%  \\
    Proposed concat. and grouped (Fig.~\ref{sfig:block3})  &  87.8\%  \\
    \bottomrule
    \end{tabular}
    \end{center}
    \caption{PCKh-based comparison on the MPII validation set for different methods of combining the features between the decoder and the encoder parts of the network. The results are reported on a model that contains approx. 3.4M parameters (2 stacks  with 128 features per block).}
    \label{tab:mpii_valid_connections_hg}
\end{table}

\subsection{Training}~\label{ssec:training}
For training, all images were center cropped to $256\times256$px around the labeled torso point. For LSP, since such point is not provided, we simple used the center of the tight bounding box. During training we randomly augmented the data on-the-fly  by applying random rotation (from $-30^{\circ}$ to $30^{\circ})$), scaling (from $0.75\times$ to $1.25\times$), flipping and color jittering. On MPII, we trained the models for 200 epochs using RMSprop~\cite{tieleman2012lecture} and a batch size of 24. During this time, the learning rate was varied from $2.5e-4$ to $1e-5$, dropping it at epochs 75, 100 and 150, while the weight decay was set to 0. The weights were initialized from an uniform distribution while the scaling factor $\alpha$ with $0$.  At the end of every HG stack we apply a pixel-wise MSE loss defined as:

\begin{equation}
    \label{eq:mse-loss}
    l = \frac{1}{N}\sum_{n=1}^N\sum_{ij}\norm{\hat{Y}_n(i,j)-Y_n(i,j)}^2,
\end{equation}
where $Y_n(i,j)$ and $\hat{Y}_n(i,j)$ represent the ground truth score map at location $(i,j)$ for the $n$th keypoint.

For LSP, we followed the best practices and finetuned the models pretrained on MPII for 100 epochs using the LSP training set + LSP-extended using the same learning rate as previously, except that on this occasion we dropped the learning rate every 25 epochs. The two missing points from LSP were obtained by simply interpolating between the annotated ones.

All of our models were implemented using pytorch~\cite{paszke2017automatic}.

\section{Experimental evaluation}\label{sec:evaluation}
In this section, we thoroughly experiment on all the parameters of our model, validate our claims and compare to the state-of-art.

\subsection{Datasets}
We perform experiments on the two most challenging datasets for single person human-pose estimation: MPII and LSP. 

\paragraph{MPII}~\cite{andriluka14cvpr} is one the most challenging datasets available to-date for articulated human pose estimation. The dataset consists of 25,000 images containing more than 40,000 annotated persons with up to 16 keypoints and occlusion labels. The images portrait humans across a large set of activities and natural scenarios collected from youtube. Out of this, and following~\cite{tompson2015efficient}, 25,000 persons were used for training and 3,000 for validation.

\paragraph{LSP}~\cite{Johnson10} is a single person human pose estimation dataset consisting of 2000 images, equally split between training and validation that contains humans performing various sport activities. Each image is annotated with up to 14 keypoints. The dataset was later expanded in~\cite{Johnson11} with 10,000 more images for training (LSP-Extended). However, many of these annotations were noisy and were re-annotated in~\cite{pishchulin16cvpr}. 

\begin{table*}[ht]
   \begin{center}
   \resizebox{0.95\linewidth}{!}{%
    \begin{tabular}{lccccccccc}
    \toprule
         \textbf{ Method} & \textbf{Head} & \textbf{Shoulder} & \textbf{Elbow} & \textbf{Wrist} & \textbf{Hip} & \textbf{Knee}  & \textbf{Ankle} & \textbf{Total} \\
          \midrule
    Pishchulin et al., CVPR'16~\cite{pishchulin16cvpr}& 94.1  & 90.2  & 83.4  & 77.3  & 82.6  & 75.7 & 68.6 & 82.4 \\
    Lifshitz et al., ECCV'16~\cite{lifshitz2016human}& 97.8  & 93.3  & 85.7  & 80.4  & 85.3  & 76.6 & 70.2 & 85.0 \\
    Gkioxary et al., ECCV'16& 96.2  & 93.1  & 86.7  & 82.1  & 85.2  & 81.4 & 74.1 & 86.1 \\
    Rafi et al., BMVC'16~\cite{rafi2016efficient}& 97.2  & 93.9  & 86.4  & 81.3  & 86.8  & 80.6 & 73.4 & 86.3 \\
    Belagiannis\&Zisserman, FG'17~\cite{belagiannis2017recurrent}& 97.7  & 95.0  & 88.2  & 83.0  & 87.9  & 82.6 & 78.4 & 88.1 \\
    Insafutdinov et al., ECCV'16~\cite{insafutdinov16ariv}& 96.8  & 95.2  & 89.3  & 84.4  & 88.4  & 83.4 & 78.0 & 88.5 \\
    Wei et al., CVPR'16~\cite{wei2016convolutional}& 97.8  & 95.0  & 88.7  & 84.0  & 88.4  & 82.8 & 79.4 & 88.5 \\
    Bulat\&Tzimiropoulos, ECCV'16~\cite{bulat2016human}& 97.9  & 95.1  & 89.9  & 85.3  & 89.4  & 85.7 & 81.7 & 89.7\\
    Newell et al., ECCV'16~\cite{newell2016stacked}& 98.2  & 96.3  & 91.2  & 87.1  & 90.1  & 87.4 & 83.6 & 90.9\\
    Ning et al., TMM'18\cite{ning2018knowledge}& 98.1  & 96.3  & 92.2  & 87.8  & 90.6  & 87.6 & 82.7 & 91.2\\
    Chu et al., CVPR'17~\cite{chu2017multi}& 98.5  & 96.3  & 91.9  & 88.1  & 90.6  & 88.0 & 85.0 & 91.5 \\
    Chou et al., arXiv'17~\cite{chou2017self}& 98.2  & 96.8  & 92.2  & 88.0  & 91.3  & 89.1 & 84.9 & 91.8 \\
    Chen et al., ICCV'17~\cite{chen2017adversarial}& 98.1  & 96.5  & 92.5  & 88.5  & 90.2  & \textbf{89.6} & 86.0 & 91.9 \\
    Yang et al., ICCV'17~\cite{yang2017learning}& 98.5  & 96.7  & 92.5  & 88.7  & 91.1  & 88.6 & 86.0 & 92.0 \\
    Ke et al., ECCV'18~\cite{ke2018multi}& 98.5  & 96.8  & 92.7  & 88.4  & 90.6  & 89.4 & \textbf{86.3} & 92.1\\
    Tang et al., ECCV'18~\cite{tang2018deeply}& 98.4  & 96.9  & 92.6  & 88.7  & \textbf{91.8}  & 89.4 & 86.2 & 92.3 \\
    Chen et al., TPAMI'19~\cite{chen2019adversarial} & 98.1  & 96.5  & 92.5  & 88.5  & 90.2  & 89.6 & 86.0 & 91.9 \\
    Ryou et al., CVPR'19~\cite{ryou2019anchor} & 98.6  & 96.6  & 92.3  & 87.8  & 90.8  & 88.8 & 86.0 & 91.9 \\
    Sun et al., CVPR'19~\cite{sun2019deep} & 98.6  & 96.9  & 92.8  & 89.0  & 91.5  & 89.0 & 85.7 & 92.3 \\
    \midrule 
    Ours (\emph{small})& 98.5  & 96.4  & 91.5  & 87.2  & 90.7  & 86.9 & 83.6 & 91.1 \\
    Ours & 98.6  & 97.0  & 93.0  & 89.2  & 91.7  & 88.9 & 86.0 & 92.4\\
    Ours* & \textbf{98.8}  & \textbf{97.5}  & \textbf{94.4}  & \textbf{91.2}  & \textbf{93.2}  & \textbf{92.2} & \textbf{89.3} & \textbf{94.1}\\
    \bottomrule
    \end{tabular}
    }
    \end{center}
    \caption{\textbf{PCKh-based comparison with state-of-the-art on the MPII test set.}  Notice that our method matches and surpasses the performance of the next top-performing method. Ours* was pre-trained on the HSSK dataset.}
    \label{tab:mpii_results}
\end{table*}

\begin{table*}[htb]
  \begin{center}
    \resizebox{0.95\linewidth}{!}{%
    \begin{tabular}{lccccccccc}
    \toprule
         \textbf{ Method} & \textbf{Head} & \textbf{Shoulder} & \textbf{Elbow} & \textbf{Wrist} & \textbf{Hip} & \textbf{Knee}  & \textbf{Ankle} & \textbf{Total} \\
          \midrule
            Yang et al., CVPR'16~\cite{yang2016end}& 90.6  & 78.1  & 73.8  & 68.8  & 74.8  & 69.9 & 58.9 & 73.6 \\
            Rafi et al., BMVC'16~\cite{rafi2016efficient}& 95.8  & 86.2  & 79.3  & 75.0  & 86.6  & 83.8 & 79.8 & 83.8 \\
            Yu et al., ECCV'16~\cite{yu2016deep}& 87.2  & 88.2  & 82.4  & 76.3  & 91.4  & 85.8 & 78.7 & 84.3 \\
            Belagiannis\&Zisserman, FG'17~\cite{belagiannis2017recurrent}& 95.2  & 89.0  & 81.5  & 77.0  & 83.7  & 87.0 & 82.8 & 85.2 \\
            Lifshitz et al., ECCV'16~\cite{lifshitz2016human}& 96.8  & 89.0  & 82.7  & 79.1  & 90.9  & 86.0 & 82.5 & 86.7 \\
            Pishchulin et al., CVPR'16~\cite{pishchulin16cvpr}& 97.0  & 91.0  & 83.8  & 78.1  & 91.0  & 86.7 & 82.0 & 87.1 \\
            Insafutdinov et al., ECCV'16~\cite{insafutdinov16ariv}& 97.4  & 92.7  & 87.5  & 84.4  & 91.5  & 89.9 & 87.2 & 90.1 \\
            Wei et al., CVPR'16~\cite{wei2016convolutional}& 97.8  & 92.5  & 87.0  & 83.9  & 91.5  & 90.8 & 89.9 & 90.5 \\
            Bulat\&Tzimiropoulos, ECCV'16~\cite{bulat2016human}& 97.2  & 92.1  & 88.1  & 85.2  & 92.2  & 91.4 & 88.7 & 90.7 \\
            Chu et al., CVPR'17~\cite{chu2017multi}& 98.1 & 93.7 & 89.3 & 86.9 & 93.4 & 94.0 & 92.5 & 92.6 \\
            Yang et al., ICCV'17~\cite{yang2017learning}& 98.3 & 94.5 & 92.2 & 88.9 & 94.7 & 95.0 & 93.7 & 93.9 \\
            Ning et al., TMM'18~\cite{ning2018knowledge}& 98.2 & 94.4 & 91.8 & 89.3 & 94.7 & 95.0 & 93.5 & 93.9 \\
            Chou et al., CVPR-W'17~\cite{chou2017self}& 98.2 & 94.9 & 92.2 & 89.5 & 94.2 & 95.0 & 94.1 & 94.0 \\
            \midrule
            \textbf{Ours} & \textbf{98.7} & \textbf{95.7} & \textbf{93.1} & \textbf{90.3} & \textbf{95.8} & \textbf{95.6} & \textbf{94.8} & \textbf{94.8} \\
            \bottomrule
    \end{tabular}
    }
    \end{center}
    \caption{\textbf{PCK-based comparison with state-of-the-art on the LSP test set.}}
    \label{tab:lsp_results}
\end{table*}

\begin{table*}[htb]
   \begin{center}
      \resizebox{1\linewidth}{!}{%
    \begin{tabular}{lcccccccc}
    \toprule
         \textbf{ Method} &  \textbf{Yang et al.~\cite{yang2017learning}} & \textbf{Wei et al.~\cite{wei2016convolutional}} & \textbf{Bulat\&Tzimiropoulos~\cite{bulat2016human}} & \textbf{Chu et al.~\cite{chu2017multi}} & \textbf{Newell et al.~\cite{newell2016stacked}} & \textbf{Tang et al~\cite{tang2018cu}}  &  \textbf{Ours} \\
          \midrule
            \textbf{\# parameters} & 28M  & 29.7M  & 58.1M  & 58.1M  & 25.5M  & 10.1M & \textbf{8.5M} \\
            \midrule 
            \textbf{PCKh} & 92.0\% & 88.5\% & 89.7\% & 91.5\% & 90.9\% & 90.8\% & \textbf{91.1}\%  \\
            \bottomrule
    \end{tabular}
    }
    \end{center}
    \caption{\textbf{Comparison in terms of number of parameters and PCKh accuracy with state-of-the-art on MPII testing set.}
    }
    \label{tab:sota_size}
\end{table*}
\subsection{Comparison with state-of-the-art}

Herein, we compare the performance of the proposed approach against that of other state-of-the-art methods on the MPII and LSP datasets. Despite being significantly shallower (4 stacks vs 8~\cite{newell2016stacked,tang2018cu}) and computationally lighter (see Table~\ref{tab:sota_size}), our model achieves top performance, surpassing many larger and heavier models. Furthermore, a wider version of it, where we simple increase the number of channels from 144 to 256 while keeping the number of stacks equal to 4, reaches state-of-the-art results on both MPII and LSP datasets.

As the results from Table~\ref{tab:mpii_results} show, our smaller model (8.5M parameters, $9.9\times10^9$ FLOPs, 4 stacks) surpasses the HG baseline~\cite{newell2016stacked}(25.5M parameters, $40\times10^9$ FLOPs, 8 stacks) and the recent efficient architecture proposed in~\cite{tang2018cu}(10.1M, $12.7\times10^9$ FLOPs, 8 stacks) while being less computationally demanding and having a smaller memory footprint. Furthermore, a wider version of our model(26M parameters, $33.5\times10^9$ FLOPs, 4 stacks) sets a new state-of-the art results, improving upon the previously best results by up to 0.5\% on certain categories. It is important to note that even the larger version of our model is computationally comparable to the original HG network ($33.5\times10^9$ vs $28.4\times10^9$ FLOPs) despite being 1.5\% better. Similarly, on LSP, our method achieves state-of-the-art results (see supplimentary material for numerical results). 

In addition to this, in order to asses the efficiency of our method  across the whole spectrum of computational resources we compare our approach against that of CU-Net~\cite{tang2018cu}. As the results from Table~\ref{tab:mpii_valid_comparison} show, our method matches or outperforms that of CU-Net in the high speed regime (0.7-4 GFlops). Since we didn't reduce the number of features in the base blocks that precedes the first HourGlass stack as in~\cite{tang2018cu}, our method performs similarly with CU-Net when only 0.5M parameters are used. As we move into the \textit{medium speed regime} the improvements offered by the proposed method become significantly larger, our approach being able to match the performance offered by CU-Net using 50\% less HG stacks, 20\% less parameters and FLOPs (see Table~\ref{tab:mpii_valid_normal}).

\section{Conclusion}\label{sec:conclusion}

In this paper, we revisited residual units and introduced a new learnable soft gated skip connections. Specifically, our proposed block has gated per-channel skip connections, where each channel has a learnable parameter that controls the data flow between the current and previous residual module. 
In addition, we introduce a hybrid network  that combines the HG and U-Net architectures which minimizes the number of identity connections within the network and increases the performance for the same parameter budget.
We demonstrate superior performance and efficiency on the challenging task of human body-pose estimation. 
Specifically, our model obtains state-of-the-art results on the MPII and LSP datasets.  In addition, with a reduction of 65\% in model size and complexity, we show no decrease in performance when compared to the original HG~network.

{\small
\bibliographystyle{ieee}
\bibliography{main}
}

\end{document}